\documentclass[conference]{IEEEtran}
\IEEEoverridecommandlockouts
\usepackage{cite}
\usepackage{amsmath,amssymb,amsfonts}
\usepackage{algorithmic}
\usepackage{graphicx}
\usepackage{textcomp}

\usepackage{booktabs}
\usepackage{caption}
\usepackage{multirow}
\usepackage{enumitem}
\usepackage{pifont}
\newcommand{\cmark}{\ding{51}}%
\newcommand{\xmark}{\ding{55}}%
\usepackage{hyperref}

\usepackage{times}
\usepackage{latexsym}

\usepackage[T1]{fontenc}

\usepackage[utf8]{inputenc}
\usepackage{microtype}

\usepackage{inconsolata}

\usepackage{graphicx}

\usepackage[utf8]{inputenc}
\graphicspath{{figs/}}

\usepackage{bm}
\usepackage{amsmath}
\usepackage{amssymb}

\usepackage{bbding}

\usepackage{booktabs}
\usepackage{multicol,multirow}
\usepackage{makecell}
\usepackage{tikz}
\usepackage[edges]{forest}
\usepackage{pgfplots}
\usepackage{pifont}

\usepackage{xcolor}
\def\BibTeX{{\rm B\kern-.05em{\sc i\kern-.025em b}\kern-.08em
    T\kern-.1667em\lower.7ex\hbox{E}\kern-.125emX}}
\begin{document}

\title{HLV-1K: A Large-scale Hour-Long Video Benchmark for Time-Specific Long Video Understanding}


\author{Heqing Zou\textsuperscript{\ddag\S}*, Tianze Luo\textsuperscript{\ddag}*,  Guiyang Xie\textsuperscript{\ddag}, Victor (Xiao Jie) Zhang\textsuperscript{\ddag}, Fengmao Lv\textsuperscript{$\flat$}, \\ \text{Guangcong Wang\textsuperscript{$\natural$}, Junyang Chen\textsuperscript{\pounds},  
Zhuochen Wang\textsuperscript{\ddag}, 
Hansheng Zhang\textsuperscript{\ddag}, Huaijian Zhang\textsuperscript{\ddag}} \\
\textsuperscript{\ddag}ByteDance  \textsuperscript{\S}Nanyang Technological University \textsuperscript{$\flat$}Southwest Jiaotong University \\ \textsuperscript{$\natural$}Great Bay University \textsuperscript{\pounds}Shenzhen University\\
  }


\maketitle
\renewcommand{\thefootnote}{\fnsymbol{footnote}}
\footnotetext[1]{Equal contributions.}
\renewcommand{\thefootnote}{\arabic{footnote}}
\begin{abstract}
Multimodal large language models have become a popular topic in deep visual understanding due to many promising real-world applications. However, hour-long video understanding, spanning over one hour and containing tens of thousands of visual frames, remains under-explored because of 1) challenging long-term video analyses, 2) inefficient large-model approaches, and 3) lack of large-scale benchmark datasets. Among them, in this paper, we focus on building a large-scale hour-long long video benchmark, HLV-1K \footnote{https://github.com/Vincent-ZHQ/HLV-1K}, designed to evaluate long video understanding models. HLV-1K comprises 1009 hour-long videos with 14,847 high-quality question answering (QA) and multi-choice question asnwering (MCQA) pairs with time-aware query and diverse annotations, covering frame-level, within-event-level, cross-event-level, and long-term reasoning tasks. We evaluate our benchmark using existing state-of-the-art methods and demonstrate its value for testing deep long video understanding capabilities at different levels and for various tasks. This includes promoting future long video understanding tasks at a granular level, such as deep understanding of long live videos, meeting recordings, and movies.
\end{abstract}


\section{Introduction}
\label{sec:intro}

The advent of large language models, particularly multimodal large language models (MM-LLMs), has revolutionized deep visual understanding \cite{zhang-etal-2024-mm, li2024llava}. These models excel in tasks such as image captioning \cite{wu2023next}, visual question answering \cite{wang2024qwen2}, and video summarization \cite{ren2024timechat,argaw2024scaling}. Long videos, spanning over an hour and containing tens of thousands of frames, pose unique challenges, including maintaining long-term dependencies \cite{wu2019long}, managing complex temporal dynamics \cite{wu2024longvideobench}, and processing vast amounts of visual information \cite{song2024moviechat}. Understanding long videos remains difficult due to: 1) the inherent complexity of extended video content \cite{zou2024seconds}, 2) the lack of efficient models for hour-level tokens, and 3) the absence of large-scale benchmark datasets.

Despite the advancements in MM-LLMs, the specific challenges posed by long videos necessitate specialized benchmarks. The sheer length and complexity of long videos introduce issues such as noise and redundancy, memory and computation constraints, and the need for effective temporal reasoning \cite{chandrasegaran2024hourvideo, zou2024seconds}. Existing benchmarks \cite{jang2017tgif, wu2024longvideobench} often fall short in addressing these challenges comprehensively, focusing primarily on shorter video clips or lacking detailed temporal annotations. This gap highlights the necessity for a dedicated benchmark that can rigorously evaluate the capabilities of models in understanding long videos, ensuring they can handle the intricacies of extended video content. 

\begin{figure*}[htbp]
\centerline{\includegraphics[width=0.85\linewidth]{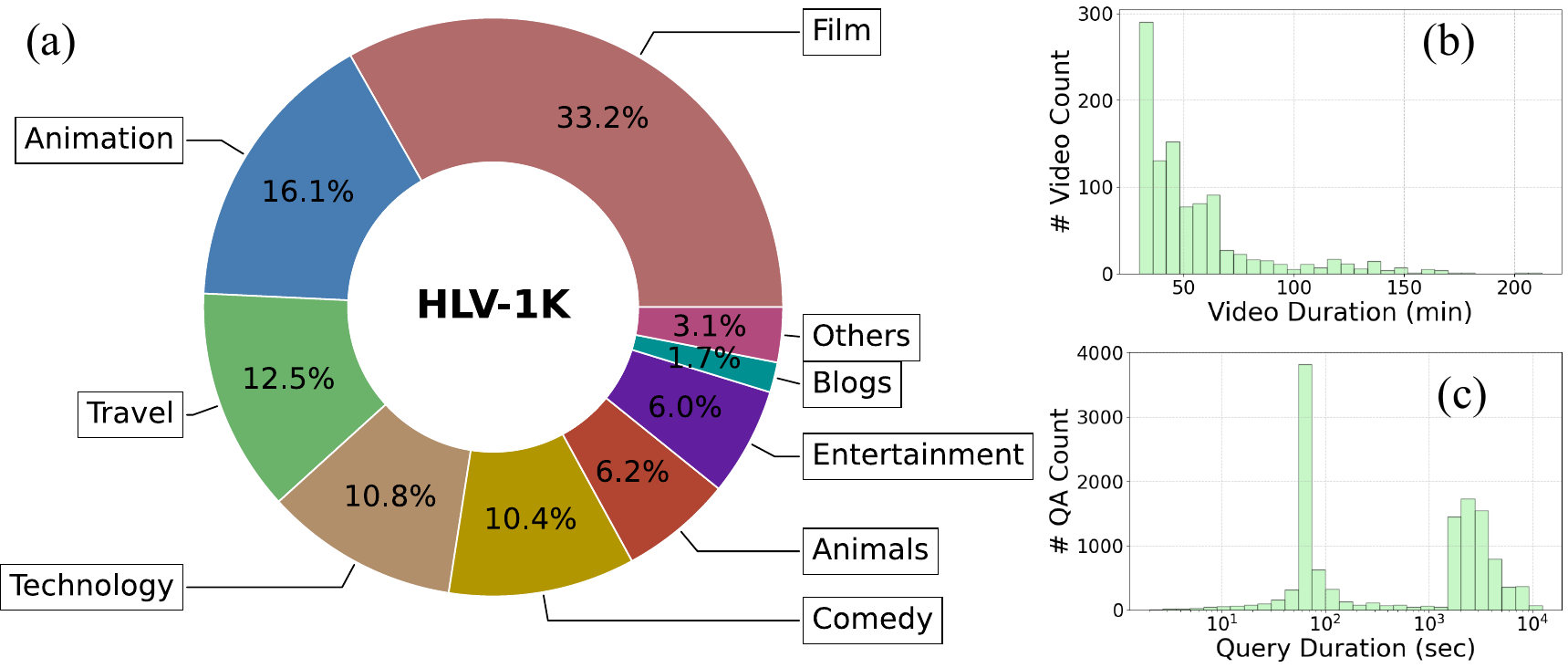}}
\caption{HLV-1K statistics: (a) Video category distribution, (b) Video duration distribution, and (c) Duration distribution of time-specific queries.}
\label{fig01}
\end{figure*}

To advance the field of long-video understanding, we introduce HLV-1K, a large-scale benchmark specifically designed to evaluate models on hour-long videos. As shown in Fig. \ref{fig01}, HLV-1K includes over 1,000 hour-long videos, annotated with 20,000 high-quality general question answering (QA) and multiple-choice question answering (MCQA) pairs with time-aware queries. As shown in Fig. \ref{fig01}(c), the query lengths vary, mapping different information in the long videos. These annotations encompass a wide range of tasks, including frame-level, within-event-level, cross-event-level, and long-term-level reasoning. The benchmark aims to provide a comprehensive evaluation framework that challenges models to maintain long-term dependencies, understand intricate temporal relationships, and process extensive visual information.

The creation of HLV-1K involves a rigorous selection and annotation process to ensure the inclusion of diverse and high-quality content. Compared with existing long video benchmarks in Table \ref{tab_dataset}, our HLV-1K has the following key features:
\begin{itemize}[leftmargin=*,labelsep=5mm]
\item \textbf{Hour-long video benchmark.} While most video benchmarks, such as MSVD-QA \cite{xu2017video}, VideoVista \cite{li2024videovista} and LongVideoBench \cite{wu2024longvideobench}, focus on second-level or minute-level video understanding, the videos in the HLV-1K dataset are all over half an hour in length.
\item \textbf{Diverse reasoning tasks.} HLV-1K includes both QA and MCQA tasks, unlike Video-MME \cite{fu2024video} and LVBench \cite{wang2024lvbench}, which mainly use MCQA. It also introduces reasoning across frame-level, within-event, cross-event, and long-term contexts for comprehensive evaluation.
\item \textbf{Time-specific queries.} Each query in HLV-1K is linked to precise timestamps, enabling robust assessment of models' temporal reasoning and long-term understanding.
\end{itemize}

To the best of our knowledge, HLV-1K is the largest long-video (i.e., each exceeds half an hour) benchmark dataset. In this paper, we detail the design and construction of the HLV-1K benchmark, highlighting the benchmark construction process. We also evaluate the dataset using existing state-of-the-art methods, demonstrating its value in advancing long-video understanding tasks. Our results reveal the strengths and limitations of current models, providing valuable insights into areas that require further research and development. By introducing HLV-1K, we aim to spur progress in the field of long-video understanding and facilitate the development of more advanced multimodal models capable of handling the complexities of long video content.

\section{Related Work}
\subsection{MM-LLMs for Long Video Understanding}
Multimodal LLMs integrate LLMs with visual encoders and excel in image and short video understanding \cite{dai2023instructblip, liu2024visual, zhang-etal-2023-video, cheng2024videollama}. However, long video understanding remains challenging due to increased spatial-temporal complexity and long-term correlations \cite{zou2024seconds}. Existing methods either compress sequential frame tokens for minute-level videos \cite{xu2024slowfast, xu2024pllava} or fine-tune on longer video datasets like Moment-10M \cite{qianmomentor}, leveraging efficient compression \cite{zhang2024longva} or long-term information preservation \cite{Ren_2024_CVPR} to enhance hour-level video understanding. These approaches aim to balance visual detail retention with computational efficiency.

\begin{table*}
  \caption{Comparison of long video understanding benchmarks}
  \centering
  \resizebox{0.95\textwidth}{!}{
  \begin{tabular}{l|ccccccc}
    \toprule
        Benchmark  & Num. of Videos & Video Duration & Num. of Labels  & Multi-Task & Multi-Level & Multi-Type & Time-Aware \\
    \midrule
    LongVideoBench \cite{wu2024longvideobench}   & 3,763   & 7.9 min  &  6,678 & {\cmark}  &   {\xmark} & {\xmark} & {\xmark}  \\
    MLVU \cite{zhou2024mlvu}  & 1,334 & 12.0 min  &  2,593 & {\cmark}  &   {\xmark}  & {\cmark} & {\xmark}  \\
    Video-MME  \cite{fu2024video}  &  900 & 17.0 min &  2,700 & {\cmark} &  {\xmark}  & {\xmark} & {\xmark}  \\  
    HourVideo \cite{chandrasegaran2024hourvideo} & 500 & 45.7 min &  12,976 & {\cmark}  &  {\xmark} & {\cmark} & {\xmark}\\
    LVBench  \cite{wang2024lvbench}  & 103  & 68.4 min &  1,549  &  {\cmark}  & {\xmark}  & {\xmark} & {\xmark}     \\
    \midrule
    \textbf{HLV-1K}  & \textbf{1,009}  & 55.0 min & 14,847 & {\cmark} & {\cmark}  & {\cmark}  & {\cmark}\\
    \bottomrule
  \end{tabular}}
  \label{tab_dataset}
\end{table*}

\subsection{Long Video Benchmarking}

To benchmark models' abilities in long video understanding, several new long video benchmarks have been introduced recently. Some datasets feature videos lasting several minutes \cite{wu2024longvideobench, zhou2024mlvu}, with some extending over an hour \cite{fu2024video, chandrasegaran2024hourvideo, zhou2024mlvu}, significantly larger than commonly used short video benchmarks with videos shorter than one minute \cite{jang2017tgif, xu2017video}. These benchmarks are designed to evaluate models' performance in handling the increased complexity and diverse content of long videos, featuring varying topics and multiple task types.

As shown in Table \ref{tab_dataset}, compared to these pioneering hour-level datasets, our proposed HLV-1K dataset includes more videos and high-quality question-answer pairs. We have clearly constructed two tasks, QA and MCQA, at different levels of long videos, with a greater diversity of task types. Additionally, our question-answer pairs are all time-specific, introducing an accurate decomposition of the time dimension in long videos. This facilitates a better exploration of the time perception ability of large models, enabling more precise and effective long video understanding.

\begin{figure*}[htbp]
\centerline{\includegraphics[width=0.95\linewidth]{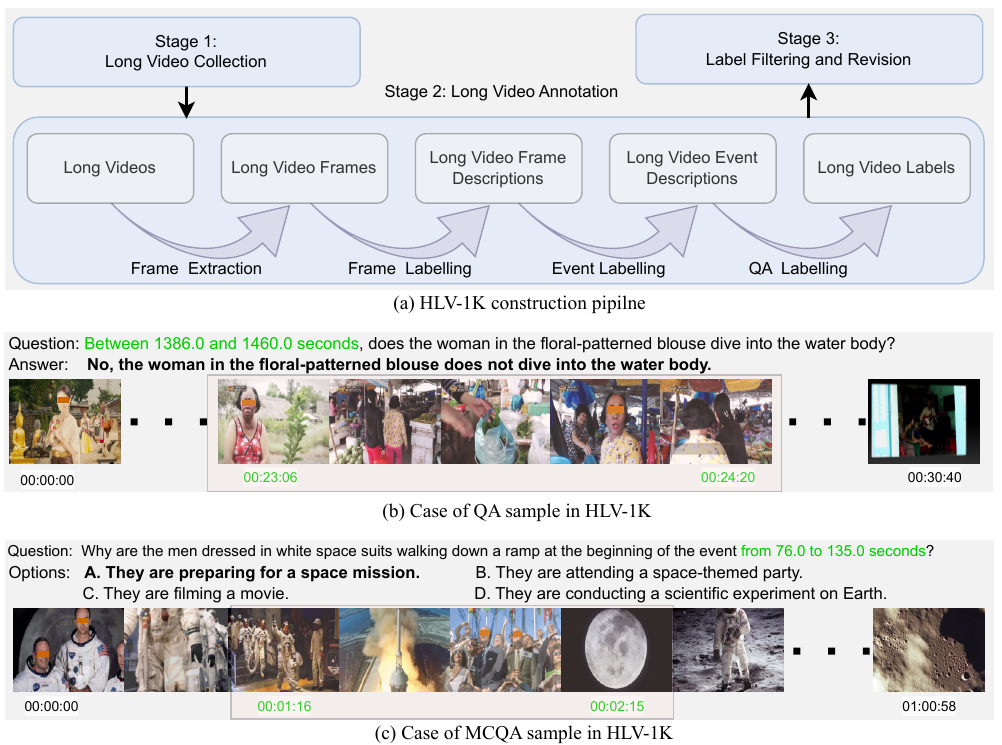}}
\caption{Construction of HLV-1K: (a) HLV-1K construction pipeline with data collection, data labeling and data filtering and revision, (b) Case of QA sample in HLV-1K and (b) Case of MCQA sample in HLV-1K.}
\label{fig02}
\end{figure*}

\section{Construction of HLV-1K}
In this section, we detail the construction of HLV-1K, including video collection, QA and MCQA labeling, and data refinement for high-quality annotations, as shown in Fig. \ref{fig02}.

\subsection{Data Collection} 
Based on the HD-VILA \cite{xue2022advancing}, a high-resolution and diverse video dataset, we collect raw videos from public resources. To obtain long videos, we select those with the largest sub-clip timestamps exceeding 30 minutes from HD-VILA and download over 1,500 long videos from YouTube. During the collection stage, we manually filter out low-quality videos and those with redundant content. Finally, we curate around 1,009 long videos covering various topics, including Entertainment, Film, Travel, Animation, Blogs, Comedy, Technology, Animals, and others such as Gaming and Music.

\subsection{Data Annotation} 
To provide high-quality long video annotations with accurate time information, we introduce a four-step labeling process to generate video QA and MCQA pairs. First, we extract dense keyframes by compressing the raw videos to obtain more content-related keyframes. Next, we label the frame descriptions using GPT-4o \cite{openai2023gpt4o}, incorporating details from YOLO \cite{yolov8_ultralytics}  and frame timestamp information. After that, we introduce a sliding-window-based event detection method and label the event descriptions with GPT-4o. Finally, we label the videos based on the corresponding frame and event descriptions to generate time-aware QAs and MCQAs at the frame-level, within-event-level, cross-event-level, and long-term-level.

\subsubsection{Frame Extraction} Raw hour-long videos typically have over 100,000 frames, containing a lot of redundant information and introducing excessive complexity to video annotation. To address this, we compress the long videos to one frame per second to reduce data redundancy. Subsequently, we extract keyframes using I-frame detection methods \cite{wiegand2003overview}. I-frames, which are the least compressible and do not require other frames for decoding, contain most of the visual information in a video. To preserve temporal continuity, we extract both the I-frame and its preceding frame. This process results in an average of 810 frames per long video.

\subsubsection{Frame Labeling} Multimodal-LLMs show remarkable performance in image understanding, and we adopt the commonly used method of annotating frame descriptions with GPT-4o. Unlike other video-annotating methods that utilize MM-LLMs directly for generating frame descriptions, we input the object position and frame time information as conditional information to GPT-4o. For object information, we use YOLO-v8 \cite{yolov8_ultralytics} for object recognition and apply a high threshold of 0.4 to identify objects and their positions in the frame images. The combination of the frame image with the object positions and frame time in the video is used to generate frame descriptions. The final descriptions contain both the spatial details and overall information of the frame, along with the frame time. The average length of the frame descriptions is over 100 words.
\begin{figure*}[htbp]
\centerline{\includegraphics[width=0.95\linewidth]{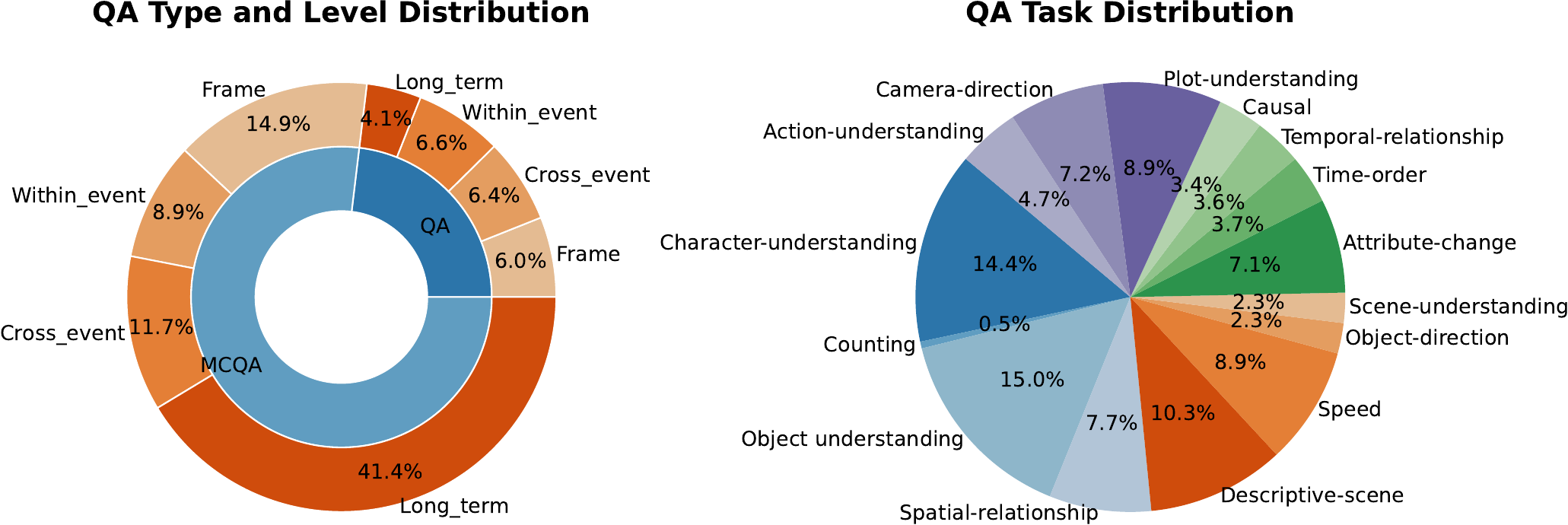}}
\caption{Distribution of benchmark annotations.}
\label{fig03}
\end{figure*}
\subsubsection{Event Labeling} The content of continuous frames is usually coherent, sharing similar information, and can be categorized into the same event. Frames from the same event share continuous visual information, while frames from different events can be significantly different. Given the limitation of inputting all frame descriptions to LLMs, we introduce a sliding-window method to generate one event description from a fixed number of frame descriptions. We set the window size to 100 frames, approximately 10 minutes of video, typically larger than most events' duration. This approach results in an average of 20 events per video, with each event description averaging around 200 words and an average event duration of 60 seconds. 

\subsubsection{QA Labeling} Based on the frame descriptions and event descriptions, we create QA and MCQA pairs as video annotations. Unlike image and short video understanding, which focus more on frame-level spatial reasoning and within-event-level spatiotemporal reasoning, long videos require additional cross-event-level and long-term-level reasoning. To align with the long video understanding task, we introduce frame-level, within-event-level, cross-event-level, and long-term-level reasoning tasks. Frame-level reasoning pairs are generated using a single frame description, while within-event reasoning pairs are derived from a single event description. In contrast, cross-event reasoning pairs are generated using two adjacent event descriptions, and long-term reasoning pairs are created using all the event descriptions from the target video. We incorporate spatial reasoning tasks for all four levels and temporal reasoning for the other three levels, except frame-level reasoning. One of our important features is preserving the temporal correspondence of the question pairs in the video. Most of the generated QA and MCQA pairs include the corresponding start and end timestamps in the video. After this step, we create around 5K QA and 15K MCQA pairs.

\subsection{Data Filtering} 
Data filtering and label revision are crucial steps to ensure the high quality of the benchmark. Initially, we filter out question-answer pairs with incorrect formats or abnormal time information. Subsequently, we manually check the annotated labels following three specific guidelines: (1) low-quality question-answer pairs, such as those with repeated options or missing-answer options, are removed directly; (2) answers that do not match the video content are revised to ensure accuracy; and (3) time information in the questions that does not align with the videos is modified to ensure a proper mapping between question-answer pairs and the video content. By adhering to these guidelines, we ultimately obtain a total of 14,847 high-quality question-answer pairs.

\subsection{Tasks of HLV-1K}
As shown in Fig. \ref{fig03}, we annotate the query labels with two types: QA and MCQA, across frame level, within-event level, cross-event level, and long-term level. The source details with varying-duration information for the multi-level annotations are summarized in the supplementary materials. The question-answer pairs cover various tasks, including character understanding, counting, object understanding, spatial relationships, descriptive scenes, speed, object direction, scene understanding, attribute change, time order, temporal relationships, causality, plot understanding, camera direction, and action understanding.

\begin{table*}
  \caption{Long video understanding evaluation results on HLV-1K with frame-level, within-event-level, cross-event level, long-term-level and the overall performance.}
  \centering
  \resizebox{0.95\textwidth}{!}{
  \begin{tabular}{l|cc|ccccc}
    \toprule
        Models  & \# LLM Params & \# Frames & Frame-level & Within-event-level & Cross-event-level & Long-term-level & Overall \\
    \midrule
    InternVL2.5 \cite{chen2024expanding} & 8B & 120 & 60.72 & 65.02 & 62.73 & 59.34 & 61.24 \\
    LongVA \cite{zhang2024longva} & 7B & 120 & 67.89 & 59.12 & 61.37 & 59.67 & 61.74 \\
    QWen2-VL \cite{wang2024qwen2} & 7B & 120 & 65.28 & 61.49 & 65.43 & 60.26 & 62.57 \\
    Kangaroo \cite{liu2024kangaroo} & 8B & 120 & 75.23 & 63.57 & 65.04 & 54.60 & 62.71 \\
    LLaVA-OneVision \cite{li2024llava} & 7B & 120 & 74.21 & 67.55 & 69.67 & 63.58 & 67.78 \\
    \midrule
    QWen2-VL \cite{wang2024qwen2} & 72B & 120 & 61.44 & 66.83 & 66.96 & 67.17 & 65.78 \\
    LLaVA-OneVision \cite{li2024llava} & 72B & 120 & 80.33 & 75.06 & 77.25 & 68.74 & 74.01 \\    
    LLaVA-Video \cite{zhang2024video} & 72B & 120 & 84.41 & 78.43 & 80.10 & 75.65 & 78.93 \\
    \midrule
    Claude 3.5 Sonnet \cite{anthropic2023claude35} & - & 20 & 26.21 & 23.98 & 27.73 & 28.89 & 27.24 \\
    GPT-4o \cite{openai2023gpt4o} & - & 120 & 53.88 & 59.08 & 56.64 & 54.37 & 55.48 \\  
    Gemini 1.5 Pro \cite{google2023gemini15pro} & - & 120 & 60.39 & 64.46 & 63.08 & 62.37 & 62.41 \\
    \bottomrule
  \end{tabular}}
  \label{tab_results}
\end{table*}

\section{Experiments}
In this section, we provide results of  the state-of-art commercial and open-source multimodal LLMs on our benchmarks. The HLV-1K dataset consists of MCQAs and QAs, with accuracies reported both per task and in aggregate across the entire dataset. A principal challenge in evaluating these MCQAs and QAs, particularly for long videos, lies in preventing information leakage across different questions. Ideally, each MCQA should be independently evaluated to mitigate such leakage. However, this independent evaluation is computationally intensive and time-consuming. Therefore, in our evaluations, we batch questions related to specific tasks or sub-tasks together. For predictive tasks (reasoning), we provide precise timestamps to trim the videos, thereby enabling targeted evaluation.

\begin{figure}[htbp]
\centerline{\includegraphics[width=1.0\linewidth]{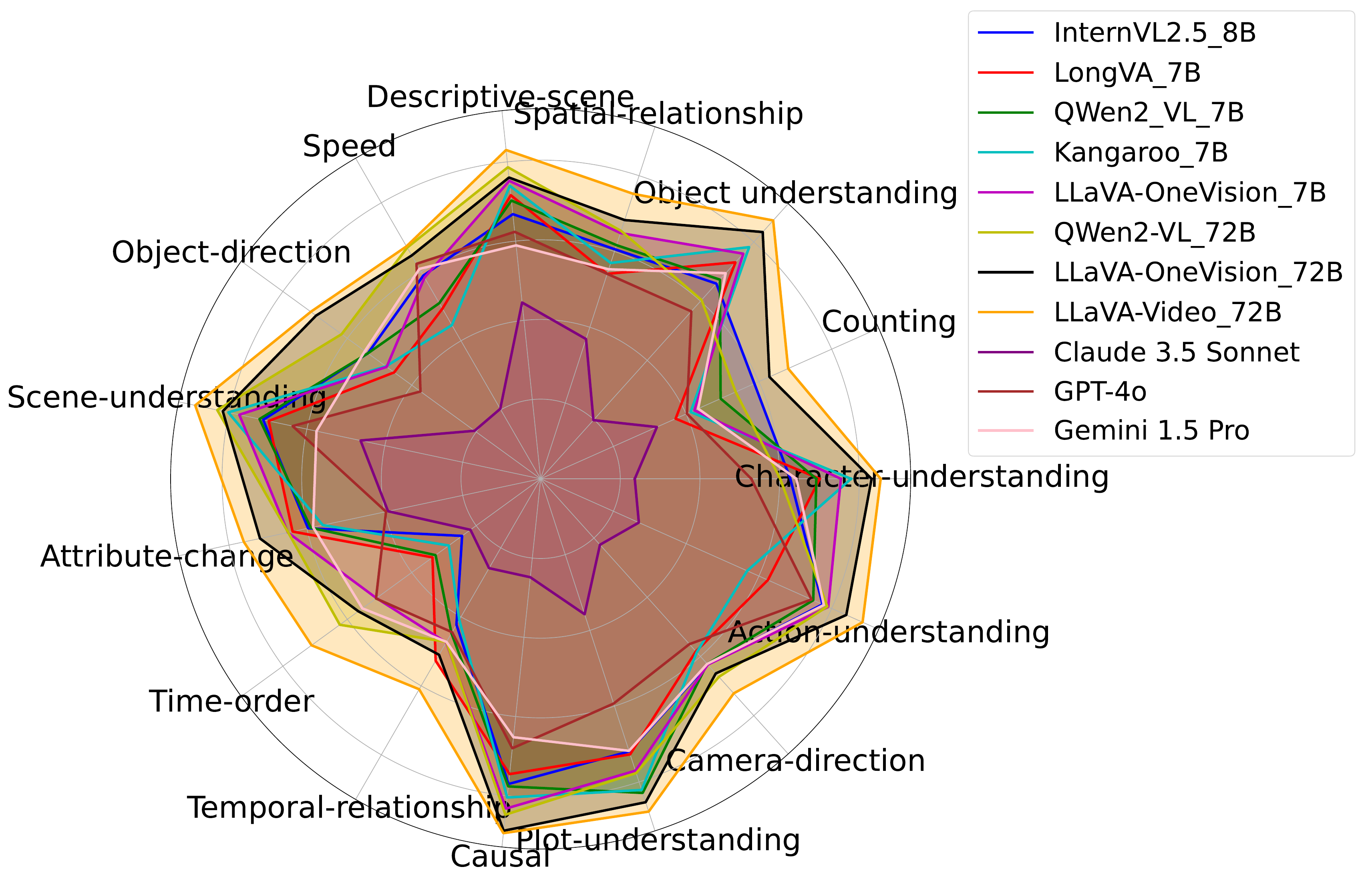}}
\caption{Long video understanding evaluation results on HLV-1K under different QA tasks.}
\label{fig03}
\end{figure}

\subsection{Settings}
We evaluate three commercial models, including GPT-4o \cite{openai2023gpt4o}, Gemini 1.5 Pro \cite{google2023gemini15pro}, and Claude 3.5 Sonnet \cite{anthropic2023claude35}, as well as state-of-the-art open-source video understanding methods, including LLaVA-OneVision \cite{li2024llava}, LLaVA-Video \cite{zhang2024video}, QWen2-VL \cite{wang2024qwen2}, Kangaroo \cite{liu2024kangaroo}, LongVA \cite{zhang2024longva}, and InternVL 2.5 \cite{chen2024expanding}, on our benchmark. We evaluate the models with a fixed frame number of 120, uniformly sampled from the video sources, except for Claude 3.5, which accepts a maximum of 20 video frames only and with same sampling method. To fairly evaluate the performance of various models, we design unified prompts for QA and MCQA tasks, respectively (details in our supplementary materials).

\subsection{Results \& Analysis}
The quantitative results for frame-level, within-event-level, cross-event-level, and long-term-level tasks, as shown in Table \ref{tab_results}, highlight several key insights. \textbf{1)} Specialized models with more parameters, such as LLaVA-Video, LLaVA-OneVision, and QWen2-VL (each with 72 billion parameters), exhibit significantly improved performance, achieving the highest overall scores. However, \textbf{2)}  larger commercial models like Gemini 1.5 Pro and GPT-4o do not perform as well in long-video understanding tasks, with Gemini 1.5 Pro scoring significantly lower at 62.41 compared to LLaVA-Video's score at 78.93. \textbf{3)} Models that accept fewer frames, such as Claude 3.5 Sonnet, which processes only 20 frames, show a marked decline in understanding and performance, underscoring the importance of frame count in video comprehension. \textbf{4)} The performance of different models varies across task levels. For frame-level tasks, LLaVA-Video and LLaVA-OneVision excel, while within-event-level and cross-event-level tasks present greater challenges, with LLaVA-Video maintaining strong performance. Long-term-level tasks, requiring deep understanding of temporal dependencies, are the most challenging, with LLaVA-Video achieving the highest score, demonstrating its superior capability in handling complex, long-term video understanding.

The radar chart in Fig. \ref{fig03} illustrates the performance of various models across different QA tasks, highlighting several key insights. \textbf{1)} Different models excel in different areas, with varying performance all tasks for ecah model. For instance, LLaVA-Video-72B and LLaVA-OneVision-72B show strong performance in tasks such as character, object, plot and scene understanding, while models like GPT-4o performs better in the speed task than some other tasks. \textbf{2)} All models struggle with tasks requiring deep understanding, such as time-order, temporal-relationship, counting, and camera-direction, indicating a general weakness in these areas. This suggests that while some models are adept at specific tasks, there is still a significant challenge in achieving comprehensive understanding across all QA tasks, particularly those involving complex temporal and spatial reasoning.

\section{Conclusion}
In this paper, we introduced HLV-1K, a large-scale long-video benchmark specifically designed to evaluate and advance the field of time-specific long video understanding. HLV-1K comprises 1009 hour-long videos, meticulously annotated with around 14,847 high-quality QA and MCQA annotations with time-aware query. These annotations span a wide range of tasks, including frame-level, within-event-level, cross-event-level, and long-term reasoning tasks, providing a comprehensive evaluation framework for multimodal large language models (MLLMs). Our benchmark addresses the unique challenges posed by long videos, such as maintaining long-term dependencies, managing complex temporal dynamics, and processing extensive visual information. Through rigorous evaluation using state-of-the-art methods, we demonstrated the value of HLV-1K in pushing the boundaries of what current models can achieve in long video understanding. We believe that HLV-1K will serve as a critical resource for the research community, fostering the development of more advanced and capable models that can effectively process and understand long video content, ultimately leading to more sophisticated applications and technologies.

\bibliographystyle{IEEEtran}
\bibliography{icme2025references}

\end{document}